\documentclass[runningheads]{llncs}
\usepackage{graphicx}
\usepackage{amsmath,amssymb} 
\usepackage{color}
\usepackage[width=122mm,left=12mm,paperwidth=146mm,height=193mm,top=12mm,paperheight=217mm]{geometry}
\usepackage{kotex}
\usepackage{subfigure}

\newif\ifdraft\drafttrue
\ifdraft

\else

\fi

\newcommand{\ie}{\textit{i}.\textit{e}. }

\begin{document}
\pagestyle{headings}
\mainmatter
\def\ECCV18SubNumber{}  

\title{Broadcasting Convolutional Network \\for Visual Relational Reasoning} 

\titlerunning{Broadcasting Convolutional Network for Visual Relational Reasoning}

\authorrunning{Simyung Chang, John Yang, SeongUk Park, Nojun Kwak}

\author{Simyung Chang$^{1,2}$, John Yang$^{1}$, SeongUk Park$^{1}$, Nojun Kwak$^{1}$ }
\institute{$^{1}$Seoul National University  \\
$^{2}$Samsung Electronics \\
\texttt{\{timelighter, yjohn, swpark0703, nojunk\}@snu.ac.kr}}

\maketitle

\begin{abstract}
In this paper, we propose the \textit{Broadcasting Convolutional Network} (BCN) that extracts key object features from the global field of an entire input image and recognizes their relationship with local features. 
BCN is a simple network module that collects effective spatial features, embeds location information and broadcasts them to the entire feature maps.
We further introduce the \textit{Multi-Relational Network} (multiRN) that improves the existing Relation Network (RN) by utilizing the BCN module. 
In pixel-based relation reasoning problems, with the help of BCN, multiRN extends the concept of `pairwise relations' in conventional RNs to `multiwise relations' by relating each object with multiple objects at once.
This yields in  $\mathcal{O}(n)$ complexity for $n$ objects, which is a vast computational gain from RNs that take $\mathcal{O}(n^2)$. 
Through experiments, multiRN has achieved a state-of-the-art performance on CLEVR dataset, which proves the usability of BCN on relation reasoning problems.
\end{abstract}

\section{Introduction}

A complete cognizance of a visual scene is achieved
by relational reasonings of a set of detected entities in an attempt to discover the underlying structure \cite{kemp2008discovery}.
Reasoning comparative relationships allows artificial intelligence to infer semantic similarities or transitive orders among objects in scenes with various perspectives and scales \cite{deformconv}.
While the core of relational reasoning instrumentally depends on spatial learning \cite{crone2009neurocognitive,raven1941standardization}, 
the relational networks (RNs) \cite{raposo2017discovering,relationalreasoningmodule} have fostered the performance vastly on related tasks based on their spatial grid features.  
However, the number of objects in conventional RNs upsurges as their method assumes that each grid represents an object at the corresponding position within the scene regardless of the existence of an object at a grid position. 
Moreover, the computational cost increases quadratically as RNs are based on pairwise computation of objects' relations for relational reasoning.

This computational burden is inevitable for visual reasoning problems if conventional architectures of convolutional neural networks (CNNs) \cite{lecun1995convolutional} are used.
Although, CNNs have allowed success in many computer vision problems \cite{girshick2014rich,krizhevsky2012imagenet,long2015fully,senet,imagenet_cvpr09},  yet
they still suffer from difficulties in generalization over geometric variations of scenes. 
This is mainly due to the receptive fields that are mapped with convolution filters at fixed areas, which derives CNNs to disregard spatial locations in the process of searching for optimal features.
Either bigger size of filters that embrace multiple input entities or 
repetitive usage of smaller filters in deeper networks are typically used to learn spatial relationships.
CNNs, however, still show limited performances for large deformations of inputs as long as the receptive fields of convolution or pooling filters stay local and small-sized \cite{jaderberg2015spatial,jeon2017active,logan1996computational,luo2016understanding}.

In order to learn relationships among objects, 
the correlation of objects needs to be defined along with the segregation of non-object features.
And, CNNs' structural loss of spatial information also needs to be overcome to handle dynamic variations of object sizes and locations.
We are motivated to solve such issues through globally extending receptive fields of object features and efficiently learning correlations among objects in an end-to-end manner.
To this end, we propose a modular technique named \textit{Broadcasting Convolutional Network} (BCN) that can be applied in any CNNs to enable learning spatial features with absolute positional information, 
broadcasting the features and analyzing visual relations among given objects. 
This technique not only overcomes the limitations of conventional narrow-sighted convolution operations by extending the receptive fields ideally to the global manner, 
but also allows to define a novel neural network called the \textit{Multi-relational Network} (multiRN) that
outperforms on the relational reasoning tasks in terms of both performance and computational efficiency. 
The proposed multiRN achieves the state-of-the-art performance on the CLEVR dataset which is the representative dataset for relational reasoning.

The paper is organized as follows. In Sections \ref{BCNsection} and \ref{multiRNsection}, the proposed BCN and multiRN are explained in detail. In Section \ref{sec:related}, the novelty of our work is described by comparing the methods with the related works. Section \ref{sec:exp} shows experimental results and finally, Section \ref{sec:conclusion} concludes the paper.

\begin{figure}[t]
\centering
\includegraphics[width=0.9\linewidth]{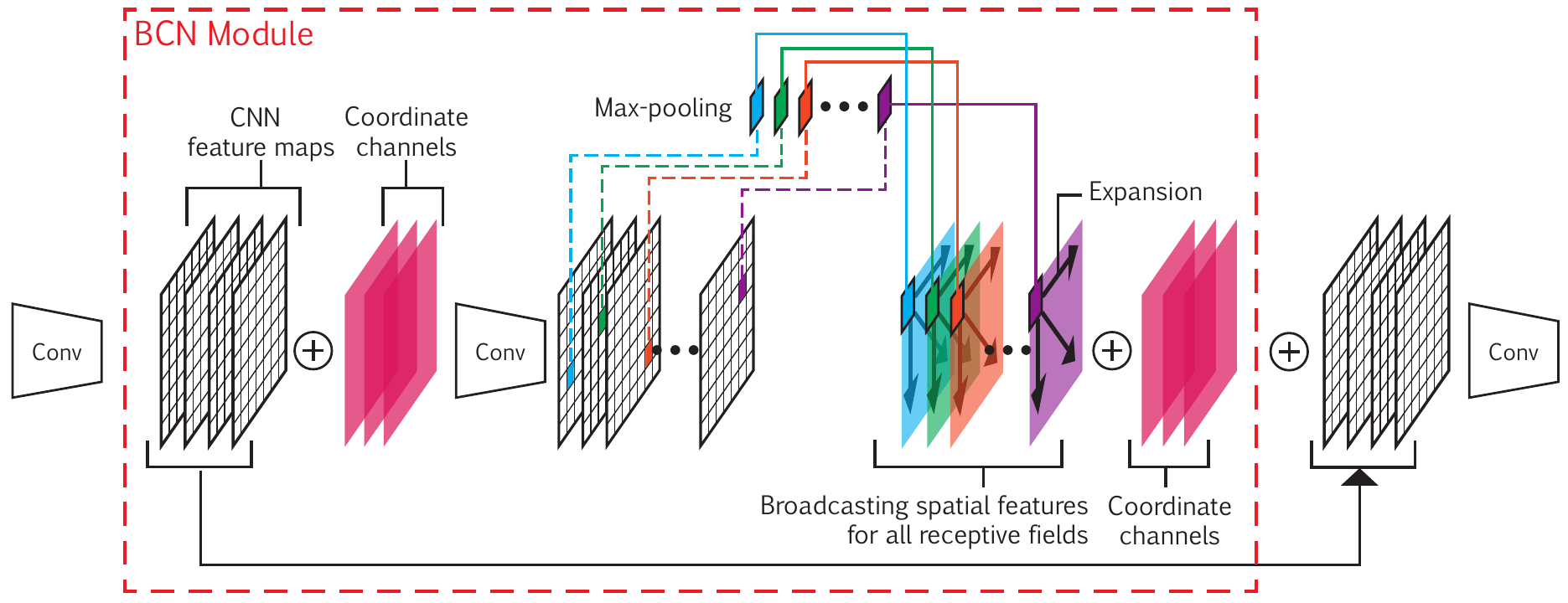}
\vskip -0.1in
\caption{\textbf{Broadcasting Convolutional Network Modules.} Feature maps acquired from previous convolution layers are concatenated with coordinate planes during CCE phase. Then, additional light-weight convolution layer(s) (\textit{e.g.} two layers of $1 \times 1$ convolution with ReLU activation) is(are) applied. An $1 \times 1 \times n$ vector is generated by max-pooling from each filter channel of the $n$ resultant feature maps. The vector is then expanded to emulate the size of the original feature maps, and merged with the original feature maps as an input to the next layer of convolution. }
\label{fig:bcn}
\end{figure}

\section{Broadcasting Convolutional Network} \label{BCNsection}
\label{sec:proposed}
In this section, we first describe the overall architecture of BCN with details of our implementations and their purposes.
The proposed BCN is depicted in Figure \ref{fig:bcn} which mainly consists of three components: 
1) coordinate channel embedding (CCE), 
2) encoding visual features with objective spatial information and
3) broadcasting globally max-pooled features through expansions.

The BCN is applied after each pixel of the feature maps acquires proper sizes of receptive fields through basic convolutions. 
The module makes feature maps that represent the coordinate information and concatenates them with the original CNN feature maps. 
Then, rather simple convolution operations (\textit{e.g.} a few layers of $1\times1$ convolution with ReLU activation) are applied, which is followed by a global max-pooling stage.
Let us say that a feature map in the shape of $h \times w \times n$ is generated from the previous convolution operations, where $n$ is the number of filters and $h$ and $w$ represent height and width of the feature maps produced, respectively. 
In this sense, a maximum element can be extracted from each filter so that an $1 \times 1 \times n$ vector is generated. 
The feature vector is then expanded to emulate the same size of the original feature maps, and then merged with the original feature maps to be convolutionally mapped together.

Our intention of such structure concentrates on reusing the relationship among current positional features and broadcasting them for global comparisons during further convolution operations.
Concatenating extracted features with the original feature maps, further convolution filters are able to correlate the objective visuospatial features (convolved features with CCE) and the relative visuospatial features (broadcasting features). 

The whole structure of BCN can be succinctly described in an equation as: 
\begin{equation}
 \mathcal{BCN}(F) = [\mathcal{E}(\max([F,C]*k), h, w), C],
\label{eq:bcn}
\end{equation}
where $[\cdot,\cdot]$ refers to the concatenation of feature maps.
Here, $F \in \mathbb{R}^{h \times w \times n'}$ is the input feature map for the broadcast convolution module, $C\in \mathbb{R}^{h \times w \times n_c}$ is $n_c$ coordinate planes, 
$*k$ represents a few layers of convolutional operations whose structure is defined in $k$ (\textit{e.g.} two layers of $1 \times 1$ convolution with ReLU activation). 
Assuming that $n$ is the number of filters in the last layer of successive convolution operations $*k$,  
the max-pooling operation is taken for each of $n$ output feature maps such that it results in an $1 \times 1 \times n$ vector. $\mathcal{E}$ denotes an expansion operation which copies its input vector to the entire $h \times w$ positions. 
The proposed BCN outputs a broadcast feature map $B \in \mathbb{R}^{h \times w \times (n+n_c)}$ which is concatenated with the original feature maps $F$ and fed to the next layer as shown in Fig. \ref{fig:bcn}. 

The convolution layer(s) right after the CCE phase can be one or multiple, 
however the major purpose of its(their) presence is to generate abstract representations based on locations.
This allows the further convolutional mappings in the later process to infer relative positions among features with the implanted information on the objective positions. 
In this paper, we have implemented two or three layers of $1 \times 1$ convolution after CCE for all experiments 
to make each pixel in feature maps being convolved depth-wise with undissolved coordinate information. 
Additionally, the number of feature maps has kept fairly small 
so that the coordinate planes can be adequately reflected to the outputs after convolution operations.
Yet, the number of coordinate planes can apparently be adjusted if needed.
This simple setting allows a large improvement on efficiency of extracting and utilizing spatial features in CNNs without much additional computation.

Since CNNs have performed generalization by taking advantage of sharing convolutional kernels in all input locations, their consequential structures are difficult to conserve spatial information throughout the layers. 
One of the intuitive ways to reflect location information to filtered outputs is to embed unique coordinates into the inputs.
In our method, the feature location information is implanted in original feature maps as additional channels.
This specific decision comes from the motivation to assign objective positional components to each feature during convolutions before the max-pooling phase, 
and to establish relative location for the comparison against other spatial features when broadcasting. 
This furthers the productivity of coordinate embedded visual features and allows additional convolutional mappings to reflect the feature positions when generating higher level features.

\begin{figure}[t]
\centering
\includegraphics[scale=0.75]{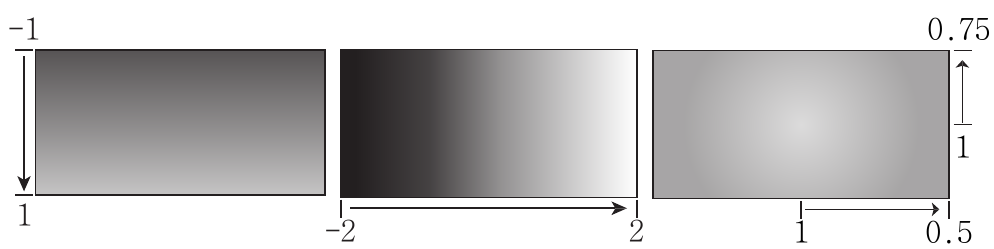}
\vskip -0.1in
\caption{An example of coordinate planes, $c_x, c_y$ and $c_r$, for feature maps with 2:1 aspect ratio of width to height. The brighter, the higher value}
\label{fig:cce_ex}
\end{figure}
Three different coordinate planes are defined as in Figure \ref{fig:cce_ex} and used in this paper; one for the x-axis, $c_x$, 
another for the y-axis, $c_y$ and the last one for the radial distance from the center, $c_r$. 
Similar to the conventional coordinate feature embedding approaches \cite{relationalreasoningmodule,darkforest,liang2015proposal,perez2017film},
these planes with the normalized coordinates reduce initial learning bias and provide additional feature location information.
Since inputs of our module may not necessarily have the shape of a square, 
elements of each plane are normalized according to the aspect ratio of inputs. 
As it can be seen in Figure \ref{fig:cce_ex}, the Cartesian coordinates $c_x$ and $c_y$ tend to be zero towards the center. This prevents features from being initially biased near bottom right-side which occurs if coordinates near top left start from zero as used in usual computer graphics. 
Also, when inputs with different aspect ratios are given, initial values in CCE are scaled to cope with the different aspect ratio.

\section{Multi-Relational Network} \label{multiRNsection}
\label{sec:multi}
\begin{figure}[t]
\centering
\includegraphics[width=0.98\linewidth]{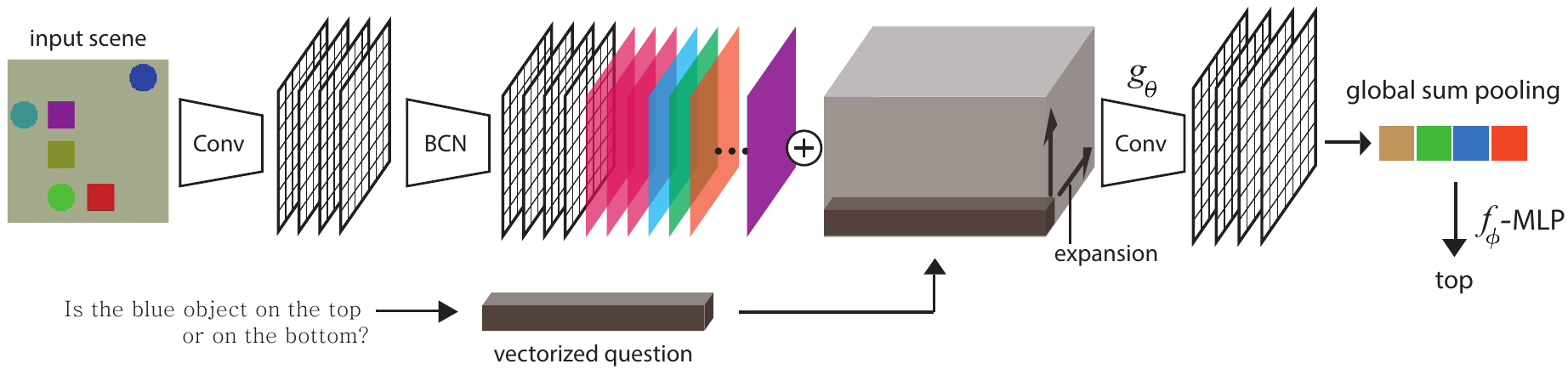}
\vskip -0.1in
\caption{\textbf{Multi-Relational Networks.} For visual relational reasoning problems, input scenes are fed into ordinary convolutional layers and a BCN module. 
Questions about the input scene are mapped to unique vectors and expanded to emulate the size of BCN outputs. 
As it can be seen above, each question vector is projected in all receptive fields.
Since the outputs of BCNs connote visuospatial features of multiple objects, they can be merged with the expanded question tensor for further convolution operations, $g_\theta$ . 
Then, global sum poolings are operated on feature maps produced, and the outcomes are integrated to answer questions }
\label{fig:mrn}
\end{figure}
Visuospatial tasks necessitate recognition of relevant features from the spatial organization of stimuli and selection of stimulus that matches one or more of these identified features \cite{crone2009neurocognitive,raven1941standardization}.
Algorithms such as relation networks (RNs) \cite{raposo2017discovering,relationalreasoningmodule} have introduced solutions for relational reasoning problems including CLEVR dataset \cite{clevr} based on their spatial grid features -- 
the dataset is explained in a more detailed manner later in this paper.
Utilizing BCN, RNs can be greatly improved in both performance and computational efficiency.

In \cite{relationalreasoningmodule}, an RN module for a set of $n$ objects, $O = \{o_1, \cdots , o_n\}$, is introduced, which consists of two functions $f_{\phi}$ and $g_{\theta}$ such that :
\begin{equation}
\mathcal{RN}(O) = f_{\phi}(\sum_i\sum_j g_{\theta}(o_i, o_j)),
\label{eq:rn}
\end{equation}
where 
$f_\phi$ represents multi-layer perceptron (MLP) operations on the visuospatial relation features 
that have been generated by another MLP operation $g_\theta$
from all pairwise combinations of objects.
The `relations' are the outputs of $\mathcal{O}(n^2)$ time of computations from $g_\theta$. 
Earlier in this paper, we have explained how BCN allows multiple spatial features to be represented in one dimensional vector.
With the help of BCN described in (\ref{eq:bcn}), original RN in (\ref{eq:rn}) can be revised into Multi-Relational Networks (multiRN) as follows:
\begin{equation}
multi\mathcal{RN}(O) = f_{\phi}(\sum_{i} g_{\theta}(o_i, \mathcal{BCN}(O))) .
\label{eq:multi_rn}
\end{equation}
Since $\mathcal{BCN}(O)$ connotes multi-location features for multiple objects, 
each object feature can be paired with multiple object location features, fed into $g_{\theta}$ and computed in $\mathcal{O}(n)$. 
This allows not only exponential gain in computation complexity but also relational comparison with multiple objects at once. 
The outputs of $g_{\theta}$ should, in our context, be redefined as `multiple relations',
and can be trained end-to-end to convey information about how much of certain object feature at particular location should be reflected for visual relational reasoning.

Furthermore, the number of objects, $n$, in pixel-based relational reasoning problems is defined by $n = h \times w$ where $h,w$ are height and width of resultant 3-dimensional tensors from previous convolution layers.
Such trait allows us to replace the MLP operation for $g_\theta$ in RNs with $1 \times 1$ convolution 
after channel concatenations of BCN outputs as shown in Figure \ref{fig:mrn}. The non-pixel-based objects such as questions in visual question and answering (VQA) problems can be additionally concatenated in channels.
The final form of our multi-relational network is : 
\begin{equation}
multiRN(O) = f_{\phi}(\sum_{i} g_{\theta}(o_i, \mathcal{BCN}(O), question).
\label{eq:multi_rn_qa}
\end{equation}
In practice, $question$s are expanded for channel embedding, 
and the final tensor is fed into $g_\theta$-convolution operations to achieve `multiple relations'.

\section{Related Works}
\label{sec:related}
Our motif has two aspects : 
1. extracting and condensing dynamic spatial features and
2. globally projecting them to previous feature maps 
which largely widen the receptive fields just as demanded in multi-scale reasonings and full resolution dense relational predictions.

Many previous computer vision related researches have used pooling techniques for extracting meaningful spatial features \cite{rfcn}.
Spatial pyramid pooling \cite{he2014spatial} uses manual control over pooling scales 
and regional features are extracted from variations of hand-engineered pooling scopes.
This issue has been developed in \cite{jia2012beyond} where a large set of various pooling bins are initiated and 
the algorithm learns to select sparse subset of them.
These works take hand-crafted pooling regions that cannot be learned end-to-end.

ROI pooling structure  \cite{deformconv,rfcn} is similar to our method's in a way of executing pooling methods in channel-wise.
Each channel represents feature maps from starting stride point to last, and thus may possess objective spatial features.
However, with conventional ROI pooling methods, each feature map does not reflect any relations among feature maps. 


While effective receptive field size is previously known to increase linearly with the number of convolutional layers \cite{luo2016understanding},
the works in \cite{deformconv} have found that it actually increases with square root of the number of layers which is a much slower rate.
This finding further leads to a logical doubt that deep CNNs may not have large enough receptive field size even at top layers. 
This phenomenon is prominent in fully convolutional networks (FCNs) with a large input image.
To overcome the issue, 
large enough receptive field size is not the only aspect that is essential but the the learning of its flexibility depending upon situations, which explains why atrous convolution methods are used widely.

Deformable convolutional neural networks (Deformable ConvNet) \cite{deformconv} is engaged with deformable convolutions that learn the applicable filters with adaptive receptive fields.
Their convolution filters are chained with offset parameters 
which represent the mapping of the original receptive fields to unique and irregularly dispersed receptive fields from each spatial location. 
The offset parameters are optimized along with the convolutional filters during back-propagation.
However, Deformable ConvNets require extra computation upon original CNNs, and their feature map sampling method in a local manner hinders themselves from having complete spatial support.
They, as their primitive motivation, further emphasize the importance of adaptive receptive learning in the needs of effectively computing large enough receptive fields at top layers.

Other atrous convolution usages include increasing the receptive field size by sampling from dilated sparse regions \cite{holschneider1990real}.
This allows to retain the same computational complexity as previous CNNs' while increasing receptive field sizes, 
and thus is widely used in semantic segmentation problems \cite{long2015fully,chen2016deeplab,yu2015multi}.
If depth conditions are excluded, the method is still doubted on utilizing enough size of receptive fields.
Our model handles extracted convolutional features by projecting them to global receptive fields, 
intending to convolve the features from various spatial locations with the original features.

Furthermore, the proposed BCN may look similar to skip-connections \cite{srivastava2015training}. 
However, in addition to the aspects of convolving features with those from previous layers, 
direct convolution operations of each feature and other broadcast visual features from various locations 
help analyzing spatial relationships in our method.
It is empirically shown in a following section that existing skip-connections improve ordinary CNNs, but not as much as  BCNs do.

A dominant approach used in the visual relational reasoning domain is the Relation Network module (RN) \cite{relationalreasoningmodule}. 
As mentioned in Section \ref{multiRNsection}, RN learns the relations among objectified features of CNN through pairwise combinations, computations of which increase quadratically with the number of the objects  
while multiRN gains a comparable  improvement on computation efficiency through manipulation of `multiple relations' induced by the BCN module.
A FiLM module~\cite{perez2017film}, as another method for relational reasoning problems, conditions features before activation functions in a similar way as done in LSTM gates~\cite{hochreiter1997long} and SENET's feature excitations~\cite{senet}. 
Such attempt of learning visual feature conditionings for reasoning largely differs from our method of extracting and broadcasting key object features to the entire receptive fields.
Also, compared to their suggested model that incorporates multiple residual blocks and large GRUs and MLPs, a multiRN model requires much smaller networks to achieve similar performances.

\section{Experiments}
\label{sec:exp}
The proposed method is tested in several experiments and compared with other similar methods to verify the effectiveness of BCN and MultiRN.
The purposes of the experiments are to investigate the followings :  

(1) the capability of BCN in representation power of features,
(2) the effectiveness of MultiRN on visual reasoning problems,
(3) feasible extensions of receptive fields caused by BCN and practical expressions of multiple objects in multiRN as specified in Section \ref{multiRNsection}, and
(4) variations of coordinate embeddings in visual features and their performances.

\subsection{Tasks}
We have experimented our methods on four datasets, Scaled-MNIST, STL-10, Sort-of-CLEVR with pixels, and CLEVR with pixels, for different purposes.

\noindent\textbf{Scaled-MNIST} dataset is our own remodeled version of original MNIST dataset \cite{mnist} to verify that the BCN can effectively handle positional information while globally extending the receptive field.
Locating the original MNIST of size $28 \times $28 to $128 \times $128 image space, 
we randomly scale the width irregularly ranging from 28 to 105 pixels and the aspect ratio from 1:0.8 to 1:1.2. 
All digits are positioned randomly within the new image space, preserving its complete form. 
Along with the class labels, another label is added for the locations of center points for each digit in order to evaluate localization performances of the models.

\noindent\textbf{STL-10} \cite{stl10} is used to evaluate the capability of BCN for natural image classification problems. 
It is a set of natural image with ten class labels each of which has 500 training images and 800 test images.  
There are also unlabeled images within the dataset, but we do not use them because experiments on them are irrelevant to our intentions. 
The data consists of images sized in 96$\times$96 with higher resolution than those of CIFAR-10 but less number of training images.

\noindent\textbf{CLEVR with pixels} \cite{clevr} is one of the visual QA datasets which is a challenging problem set requiring high-level relational reasonings. 
The dataset contains images of 3-D rendered objects and corresponding questions asking about several attributes of the objects. 
We have experimented only on the pixel version of the CLEVR dataset whose images are represented in 2-D pixel-wise.

\noindent\textbf{Sort-of-CLEVR with pixels} \cite{relationalreasoningmodule} is our main experiment for multiRN. 
Sort-of-CLEVR is a more simplified version of CLEVR, which is a set of images combined with caption dataset for relation and non-relation reasonings. 
Each image contains six of differently colored 2-D geometric shapes, and corresponding 20 questions; 10 for relational and 10 for non-relational reasonings. 
Questions in this dataset are already vectorized, and thus the experiments are independent from any additional vector embedding models, which allows more reasonable comparison based on the results.

\setlength{\tabcolsep}{4pt}
\begin{table}[t]
\caption{The model with BCN has the highest classification accuracy and the lowest localization error on the Scaled-MNIST data.
Skip-connection denotes using the shape of $1 \times $1 convolution layers for BCN. This can be viewed just the same as skip-connection \cite{srivastava2015training}}
\label{table:exp_smnist}
\centering
\vskip -0.05in
\resizebox{0.98\linewidth}{!}{
\begin{tabular}{l|c|c|c|c}
\hline  
Model & Classification Acc.  & Localization Err.& \#Params & Runtime \footnotemark[1] \\
\hline
Baseline & 84.4\% & 0.151 & 11.2K & 4.5ms \\
Baseline(depth 4) & 94.9\% & 0.149 & 16.5K & 5.1ms \\
Baseline(depth 5) & 96.4\% & 0.089& 21.9K & 5.5ms \\
Baseline(depth 5, 2$\times$ filters) & 97.2\% & 0.077& 85.2K & 8.5ms \\
\hline
Base + Deformable Conv \cite{deformconv}& 90.8\% & 0.087 & 32.0K & 16.7ms \footnotemark[2] \\
Base + Dilated Conv \cite{yu2015multi}& 91.0\% & 0.152 & 11.2K & 4.7ms\\
\hline
Base + CCE & 84.8\% & 0.088 & 11.9K & 4.6ms \\
Base + Skip-connection & 87.0\% & 0.151 & 29.0K & 8.4ms \\
Base + CCE + Skip-connection & 87.9\% & 0.071 & 29.2K & 8.41ms \\
Base + BCN w/ average pooling & 92.7\% & 0.064 & 29.2K & 8.42ms \\
\hline
\textbf{Base + BCN} & \textbf{97.5}\% & \textbf{0.023} & 29.2K & 8.42ms \\
\hline
\end{tabular}} 
\end{table}
\setlength{\tabcolsep}{1.4pt}

\subsection{Evaluation of BCN}
For all experiments, BCN uses multiple convolution layers with different size of $1\times 1$ kernels and ReLUs for non-linearity. The number of filters in $m$ convolution layers for a BCN is written in the form of $[S_1, \cdots, S_m]$ in each experiment.

\noindent\textbf{Scaled-MNIST:}
As a baseline model, we have stacked three to five of convolutional layers with 24 of $3 \times 3$ kernels, stride size of 2 and padding size of 1 for each edge, followed by a ReLU activation function and a batch normalization, depending on the comparison target. 
With the same baseline model (a 3-layered CNN), 
experiments of deformable convolutions and dilated convolutions are also done. 
The deformable convolution filters are applied to all three layers, and the dilated convolution filters are applied in the second layer of the baseline model with $2\times2$ dilation. 
A BCN with an output channel length of $[64,64,128]$ is applied once in between the second and third layers. 
After applying the BCN, the number of feature maps increases, so in order to match the input dimension of the third layer, $1 \times 1$ convolutions are operated for dimension reduction to 24. This setting is purposefully designed to compare performance of a simpler model with BCN against that of baseline model with more depths and kinds of convolution on the given data.
Also, the network structure of expanding channels from 24 to 128 is intended for having more channels when globally max-pooling spatial features to preserve enough visual context information for broadcasting. 

\footnotetext[1]{All runtimes are measured on Nvidia Titan X (Pascal) GPU and 8 core CPU(i7-6700K) per 100 samples. }
\footnotetext[2]{The experiment on deformable convolution is performed with a pytorch implementation~\cite{Wei2017deform}.}
Table \ref{table:exp_smnist} shows the performance enhancement of our method on both classification and localization results on the Scaled-MNIST dataset. 
Considering the baseline model achieves 97.8\% classification accuracy in our experiment on the original MNIST, 
its performance of the Scaled-MNIST clearly implies its structurally inherent limitation as the size of the required receptive field increases
while making the model deeper to extend the receptive field yields a better result as shown in the table.
Even if we increase the receptive field by deepening the convolution net with up to five layers, 
it can be seen that our method with a shallower depth of three layers reaches a higher performance. 
The model with BCN even shows better performance than the 5-layer baseline model with twice the number of filters by 0.3\% for classification and three times less error for localization.

Furthermore, the significance of the broadcasting phase can be well analyzed from the results.  
While embedding additional coordinate information reduces the localization error and the skip-connection alone improves the classification performance, 
compared to the model using both, the model with BCN improves the classification accuracy by 9.6\% and reduces the localization error by 0.48. 
Besides from aspects of CCE and the skip-connection technique, concatenation of expanded max-pooled features of BCN allows the network to globally broaden receptive fields and directly learn visually relational features. And the result of the BCN using average-pooling instead of max-pooling shows a significant performance decline, which indicates that average-pooling is not a suitable method considering that the purpose of max-pooling is to extract key features.

The goal of deformable convolutions is similar to ours in many ways, 
and the implementation has also led to desirable results, getting better scores on both classification and localization. 
The model with dilated convolutions has also resulted in an impressive improvement on classification accuracy 
while conserving the same parameter numbers and computation speed as those of the baseline. 
Nonetheless, the performance of BCN model outperforms both models.
In addition, BCN shows relatively high robustness against deformations compared to other methods, and the experimental results are included in the supplementary material.

\setlength{\tabcolsep}{4pt}
\begin{table}[t]
\caption{\textbf{Results on STL-10.} All models are trained from scratch without any external dataset }
\label{table:STL10_exp}
\centering
\resizebox{0.7\linewidth}{!}{
\begin{tabular}{l|c|c|c}
\hline
Model & Accuracy& \#Params & Runtime$^1$ \\
\hline
Tho17-2 Single \cite{thoma2017analysis} & 75.76\%& 1.46M & - \\
Tho17-2 Ensemble \cite{thoma2017analysis}& 78.66\%& 1.46M & - \\
\hline
Baseline & 68.75\%& 114K & 18.1ms \\
Baseline + Skip-connection & 69.79\%& 192.7K & 18.2ms \\
\textbf{Baseline + BCN} & \textbf{72.18\%}& 193.2K & 18.2ms \\
\hline
Resnet18 & 76.27\%& 11.2M & 102.2ms \\
\textbf{Resnet18 + BCN} & \textbf{77.00\%}& 11.46M& 105.5ms \\
\hline
\end{tabular}} 
\end{table}
\setlength{\tabcolsep}{1.4pt}

\noindent\textbf{STL-10 :} For a baseline model for the STL-10 dataset, 
four layers of convolutions with 64 of $3 \times 3$ kernels with the stride size of 2 and the padding size of 1 for each edge are applied. A ReLU activation function and a batch normalizing operation follow after each convolution layer. 
A BCN module of size [128, 128, 256] is applied in between the third and the fourth convolution layers. 
After applying the BCN module, the number of feature maps increases, so in order to match the input dimension of the fourth layer, an $1 \times 1$ convolution layer is used for dimension reduction to 64. 
We, in addition, have applied the same size of BCN to a Resnet18 model \cite{resnet}. 
The BCN module within the Resnet18 model is applied between the third and the fourth residual blocks. 
Then, 256 of $1 \times 1$ convolution filters are used to reduce number of feature maps, and the entire model is trained end-to-end from scratch.

We compare our model against the baseline model in Table~\ref{table:STL10_exp} to evaluate the performance of BCN for a natural image classification problem. 
Because BCN extends the receptive field size to whole image grids, 
it allows 3.4\% of performance enhancement.
Also, this is an even better result than the model using the same convolution layers of BCN as a Skip-connection.
The additional experiment where BCN is applied within the Resnet18 model shows that 
our approach sets a new state-of-the-art performance in STL-10 experiment on a single network basis with 77.0\% accuracy, which is 1.24\% higher than the single model of Tho17-2~\cite{thoma2017analysis}.

\setlength{\tabcolsep}{4pt}
\begin{table}[t]
\caption{\textbf{Results on Sort-of-CLEVR.} 
RN$^*$ is the reproduced result with same model of \cite{relationalreasoningmodule} on a single GPU. 
RN$\dagger$ is a model of which the network structure is set the same as multiRN except that  pairwise comparison for a fair comparison.
CNN$_h$ denotes the CNN that has 1 stride for fourth convolution layer instead of 2 stride to handle more objects
}
\label{table:SoC_exp}
\centering
\resizebox{0.90\linewidth}{!}{
\begin{tabular}{l|c|c|c|c}
\hline
Model & Relational & Non-relational & \#Params & Runtime$^1$\\
\hline
CNN+RN \cite{relationalreasoningmodule}& 94\%$\uparrow$ & 94\%$\uparrow$ & 19.5M & -\\
CNN+RN$^*$ & 91.0\% &99.6\% & 19.5M & 575.8ms\\ 
\hline
CNN+RN$\dagger$ & 89.9\% & 99.8\% & 365K & 23.5ms \\
CNN$_h$+RN$\dagger$ & 96.5\% & 99.9\% & 365K & 315.6ms \\
\hline
CNN+MLP & 74.2\% & 65.0\% & 239K & 6.2ms \\
CNN+CCE+MLP & 72.9\% & 64.5\% & 258K & 6.2ms\\
CNN+multiRN w/o BCN & 88.7\% & 99.3\% & 224K & 7.5ms \\
\hline
\textbf{CNN+multiRN}  & 92.9\% & 99.9\%   & 345K   & 8.3ms   \\
\textbf{CNN$_h$+multiRN} & \textbf{96.7\%} & \textbf{99.9\%}   & 345K & 9.9ms   \\
\hline 
\end{tabular}
} 
\end{table}
\setlength{\tabcolsep}{1.4pt}

\subsection{Pixel-based Relational Reasoning Problem}
\textbf{Model:} 
For relational reasoning experiments, we construct our multiRN model based on 
the RN model that is used for the CLEVR dataset out of two models reported in~\cite{relationalreasoningmodule} which consists of one trained for the Sort-of-CLEVR dataset and a shallower model for the CLEVR.
Four convolution layers with 24 of 3$\times$3 kernels, followed by ReLU activations and batch normalizations, are used for the CNN part. 
MultiRN consists of a BCN of size [128, 128, 256], two convolution layers with 256 1$\times$1 kernels for $g_\theta$, and two MLP layers of 256 units for $f_\phi$. 
Also, to verify that multiRNs use computation resource efficiently, 
we change the fourth convolution layer's stride from 2 to 1 in the CNN$_h$, 
which quadruples the number of objects that the following network has to handle. 
MultiRN, for the CLEVR task, uses LSTM of 128 hidden unit of 2 layers for natural language question processing.

\noindent\textbf{Sort-of-CLEVR:} 
Results of Table \ref{table:SoC_exp} suggest that using our version of RN (multiRN) is better than the RN with similar structure at reasoning relations by 3\%, which is the main job of Sort-of-CLEVR task, despite of comparably less computational cost.
Compared to both RN and multiRN, CNN+MLP model performs far poorly. 
Even embedding coordinate information into the same model worsens the performance, supposedly, due to overfitting. 
This implies that the performance enhancement of RN and multiRN is not simply caused by the addition of coordinate maps. 
For a further ablation study, the BCN module is removed along with the coordinate encoding function, while having $g_\theta$ and $f_\phi$ remained in order to generate the `CNN+multiRN w/o BCN' model.
This model lacks information of correlations among multiple objects, and thus performs 4.2\% lower than the CNN+multiRN model. 
To check the efficiency when the number of object inputs upsurges for the multiRN to manage, we 
reduce the stride of the last layer of the preceding CNN from 2 to 1, denoting as CNN$_h$.
Since the computation complexity of multiRN is $\mathcal{O}(n)$, which is a strong advantage over RNs with $\mathcal{O}(n^2)$ for $n$ number of objects, 
the multiRN's computation cost is only quadrupled (25 to 100) for CNN$_h$ while RN takes 16 times greater
computation loads (625 to 10000). 
Both RN and multiRN with CNN$_h$ have achieved impressive performance gain with increase in the number of feature objects, but there is a large difference in computational efficiency.

\setlength{\tabcolsep}{4pt}
\begin{table}[t]
\caption{\textbf{Results on CLEVR from Pixel.} 
RN$^*$ is the result when we reproduce the same model as the paper~\cite{relationalreasoningmodule} on a single GPU. $\ddagger$ denotes the result of changing the size of an input image from 128$\times$128 to 224$\times$224, which is the same as FiLM~\cite{perez2017film}}
\label{table:clevr_expl}
\centering
\resizebox{0.98\linewidth}{!}{
\begin{tabular}{l|c|ccccc}
\hline
Model            & Overall & Count & Exist & \begin{tabular}[c]{@{}l@{}}Compare\\Numbers\end{tabular} & \begin{tabular}[c]{@{}l@{}}Query\\Atrribute\end{tabular} & \begin{tabular}[c]{@{}l@{}}Compare\\ Attribute\end{tabular} \\ \hline
Human \cite{johnson2017inferring}  & 92.6    & 86.7  & 96.6  & 86.5   & 95.0                                                      & 96.0                                                        \\ \hline
Q-type baseline \cite{johnson2017inferring} & 41.8    & 34.6  & 50.2  & 51.0 & 36.0                                                      & 51.3                                                        \\ 
LSTM \cite{johnson2017inferring} & 46.8    & 41.7  & 61.1  & 69.8  & 36.8                                                      & 51.8                                                        \\ 
CNN+LSTM  \cite{johnson2017inferring} & 52.3    & 43.7  & 65.2  & 67.1 & 49.3                                                      & 53.0                                                        \\ 
CNN+LSTM+SA \cite{johnson2017inferring} & 68.5    & 52.2  & 71.1  & 73.5 & 85.3                                                      & 52.3                                                        \\ 
CNN+LSTM+RN \cite{relationalreasoningmodule} & 95.5 & 90.1  & 97.8  & 93.6  & 97.9  & 97.1      \\ 
CNN+LSTM+RN$^*$     & 90.9   & 86.7 & 97.4  & 90.0  &  90.2 &  93.5 \\ 
CNN+GRU+FiLM with ResNet-101~\cite{perez2017film} & \textbf{97.7}   & 94.3 & 99.1  & 96.8  &  99.1 &  99.1 \\  
CNN+GRU+FiLM from raw pixels~\cite{perez2017film} & 97.6   & 94.3 & \textbf{99.3}  & 93.4  &  \textbf{99.3} &  \textbf{99.3} \\  
\hline
\textbf{CNN+LSTM+multiRN} &92.3 & 85.2 & 96.5& 93.6 & 95.1  & 92.9\\ 
\textbf{CNN$_{h}$+LSTM+multiRN} & 97.2 & 94.1 &98.9 & \textbf{98.3} & 98.6  &  97.6 \\ 
\textbf{CNN$_{h}$+LSTM+multiRN$\ddagger$} & \textbf{97.7} & \textbf{94.9} & 99.2 & 97.2 & 98.7  & 98.3 \\ 
\hline
\end{tabular}
}
\end{table}
\setlength{\tabcolsep}{1.4pt}
\noindent\textbf{CLEVR:} An experiment has been done on a more challening relational reasoning problem to test the performance of multiRN compared to existing methods. 
The model is replicated having internal RN module replaced with multiRN to create `CNN+LSTM+MultiRN'.
The structure of multiRN model is the same as it was used for Sort-of-CLEVR dataset, and question vectors are generated from an LSTM model which is also implemented in the original RN module.

The results of the comparative experiments on CLEVR are shown in Table \ref{table:clevr_expl}. 
The RN*, our reproduced version of the relational network from \cite{relationalreasoningmodule},  has not been able to achieve the same performance as in the paper 
due to lack of original version's details of hyper-parameters and additional network control factors. 
However, since the input convolutional feature maps and question vectors are generated with the same CNN+LSTM settings, reasonable comparisons can still be made among the candidate methods. 
For this problem, our implementation using multiRN performs 1.4\% better than RN*, and multiRN with CNN$_h$ achieves even far better performance enhancement of 6.3\%. 
This is an impressive result considering that the same CNN+LSTM+RN model architecture is used as in RN* and allows 95.5\% of performance
whilst both of multiRN and multiRN with CNN$_h$ require a much smaller amount of computation. 
We have not been able to proceed an experiment with the CNN$_h$+LSTM+RN model because of the out-of-memory problem when dealing with an increased number of objects.
Furthermore, our model achieves 97.7\% of test accuracy by changing the size of an input image to 224$\times$224 which is the same as it is in FiLM~\cite{perez2017film}. For the best of our knowledge, this score is the state-of-the-art performance on CLEVR with raw pixels, and is compatible with FiLM using ResNet-101~\cite{resnet} pre-trained on ImageNet~\cite{imagenet_cvpr09}. 

\begin{figure}[t]
\centering
\includegraphics[width=.98\linewidth]{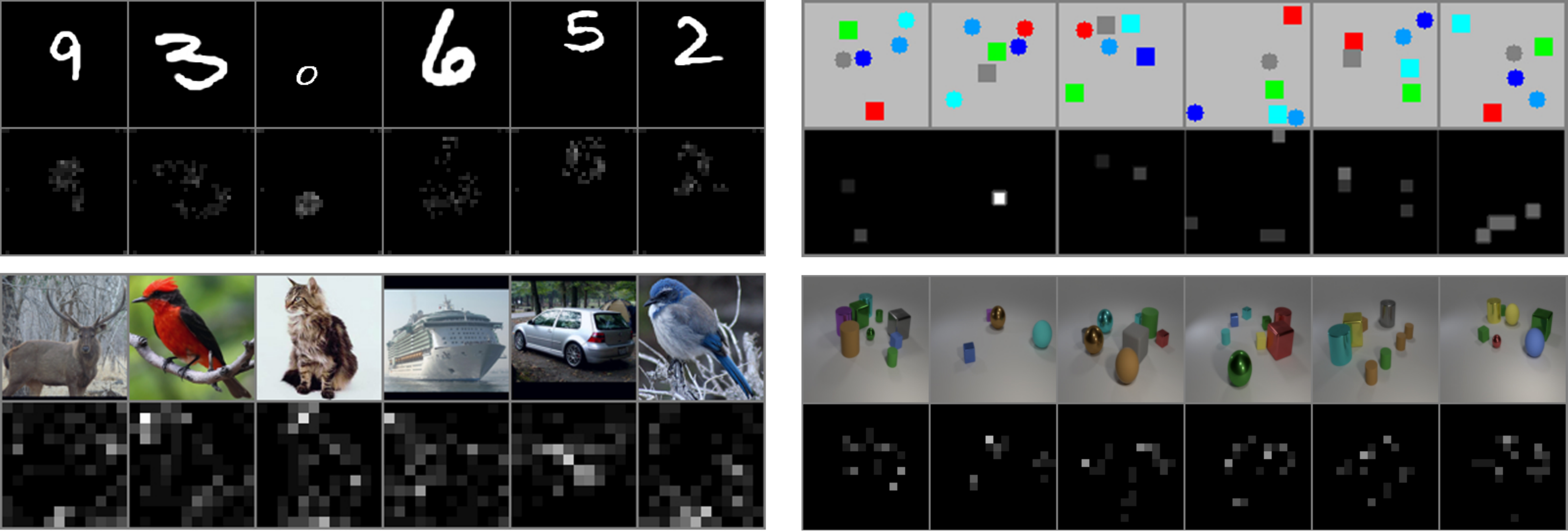}
\vskip -0.1in
\caption{\textbf{Activation map} shows how much information is broadcasted at each location.
From top to bottom are Scaled-MNIST and STL-10 on the left, Sort-of-CLEVR and CLEVR on the right. More activation maps are included in the supplementary material
}
\label{fig:act_map}  
\end{figure}
\subsection{Can BCN Extend the Receptive Field Globally and Represent Multiple Objects?}
For better understanding of our model, visualization of activation maps for different images of each experiment is provided in Figure~\ref{fig:act_map}. 
Activation maps are acquired by unsampling the global max-pooling layer, and simply masking the pixel as 1 from where it was chosen. 
The number of masks is the same as the number of $1\times 1$ convolution filters, and summing them through feature dimension outputs of the activation map. 
In activation maps for Scaled-MNIST, we can intuitively find that the activation map corresponds well with each digit's location in the original image. 
The Scaled-MNIST dataset may have too many zero inputs for a network to easily respond to spatial features, but activation maps in STL-10 apparently show that BCN makes abstractions of features well even in natural images.
Notably, we could apparently observe that activation maps tend to draw features from important locations, such as edges, faces, legs and so on. As this information is broadcast to all locations, the receptive field expands globally.
For Sort-of-CLEVR task and CLEVR task, we can obviously see that BCN is well trained to make abstractions of multiple objects in their image. This means that the `multiple relations' of multiRN described in Section \ref{multiRNsection} is established through BCN. Note that since activation maps only represent the maxpooled features, it does not have to include all objects by themselves. Output convolution features that are inputted to the BCN module, which will be further combined with BCN output, also contain features of objects.

\subsection{Study on Coordinate Embedding Methods}
\setlength{\tabcolsep}{4pt}
\begin{table}[t]
\centering
\caption{Results on coordinate embedding methods. Our method outperforms conventional coordinate methods in both Scaled-MNIST and Sort-of-CLEVR}
\label{table:cood_exp}
\resizebox{0.99\linewidth}{!}{
\begin{tabular}{l|cc|cc}
\hline
 & \multicolumn{2}{c|}{Scaled-MNIST} & \multicolumn{2}{c}{Sort-of-CLEVR} \\ \hline
Method & Accuracy & Localization err. & Relational acc. & Non-Relational acc. \\ \hline
No Coordinate channels & 92.5\% & 0.151 & 91.4\% & 91.8\% \\ 
X,Y Coordinates with top-left zero & 96.9\% & 0.029 & 92.6\% & 99.9\% \\ 
X,Y Coordinates with zero-centered & 97.5\% & 0.025 & 96.1\% & 99.9\% \\ \hline
\begin{tabular}[c]{@{}l@{}}\textbf{X,Y Coordinates with zero-centered}\\ + \textbf{Radial distance}\end{tabular} & 97.5\% & 0.023 & 96.7\% & 99.9\% \\ \hline
\end{tabular}
}
\end{table}
\setlength{\tabcolsep}{1.4pt}
\begin{figure}[t]
\centering
\includegraphics[width=.5\linewidth]{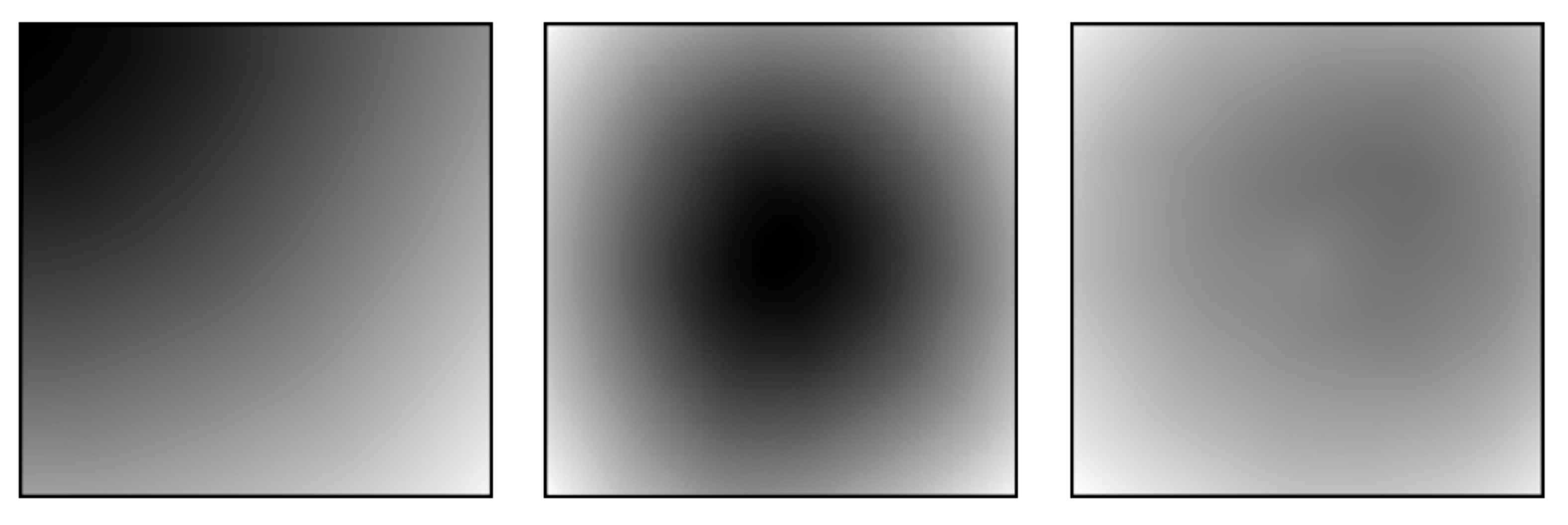}
\vskip -0.1in
\caption{The outputs of coordinate planes through the convolution kernel. Left: X,Y coordinates with top-left zero. Center: X,Y coordinates with zero-centered. Right: X,Y coordinates with zero-centered and additional Radial channel. The right has the least deflection }
\label{fig:cce_map}
\end{figure}
Our demonstration on the effects of coordinate channels is shown in Table \ref{table:cood_exp}.
The result without coordinate channels has far less score compared to the result with the coordinate channels embedded. 
The model of coordinates with zero-centered shows a significantly higher relational accuracy than the model of conventional coordinates with top-left zero for Sort-of-CLEVR, but it is slightly worse than that of three planes with extra radial distance plane included.
Figure~\ref{fig:cce_map} shows the absolute values of the output from the convolution kernel with randomly initialized weights passing through the coordinate channels.
As shown, a coordinate system with a conventional top-left of zero is deflected diagonally, but a coordinate system with zero center is deflected to the outside. 
On the other hand, by adding coordinate channels with radial distances, it can be seen that most of the deflection can be removed while providing additional coordinate information.
Furthermore, although we have expected better feature representation and localization performance by concatenating an additional coordinate plane and reducing initial weight bias, the performance gain has only occurred in the localization aspect.
This evidently implies that convolution operations are inherently biased towards the center with Gaussian distribution \cite{luo2016understanding}, and thus the additional coordinate plane have not been a critical catalyst. 

\section{Conclusion}
\label{sec:conclusion}

We have shown that utilizing Broadcasting Convolutional Network (BCN) allows conventional CNNs
to effectively collect and represent spatial information with
efficient extension of receptive fields, which results in
remarkable performance on localization problems.
With BCN's ability of representing compounded spatial features in all receptive fields, 
we have proposed Multi-Relational Networks that greatly improve RN \cite{relationalreasoningmodule}  in terms of computational gains while achieving a state-of-the-art performance in pixel-based relation reasoning problems.

In future works,
we intend to study whether BCN can be applied to other domains such as object detection or semantic segmentation. And we need to study applying multiRN to various problems that require visual relational reasoning. 

\bibliographystyle{splncs}
\bibliography{egbib}

\clearpage

\setcounter{section}{0}
\renewcommand\thesection{\Alph{section}}

\noindent{\Large Appendix}
\section{Experiment Details}
For the ease of reproduction of our experiment results, this section explains as much details as possible of our experiments. 
All our implementations of neural networks are based on Pytorch of version 0.2.0.

\subsubsection{Scaled-MNIST}
Scaled-MNIST dataset is created one time and has stayed intact
during training, \ie any further transformations {or any data augmentation techniques have {not} been applied throughout the experiments.}
The models are optimized with the stochastic gradient descent method with learning rates of 0.01 during first 10 epochs, 0.001 during next 10 epochs and 0.0001 during last 10 epochs. 
However, the network using deformable convolutions have empirically required much more sensitive control during training.
We therefore have heuristically chosen the learning rate of 0.00007 to avoid the over-fitting. 
All models are trained for 30 epochs and early-stopped because almost all of the experimented models become saturated before reaching 30 epochs. 
Our loss function is composed of two different loss functions:  
the loss function for classification that computes negative log likelihood cross-entropy with softmax outputs 
and the loss function for localization with mean square error functions for the center location of digits. 
Both of them are weighted equally in training.

\subsubsection{STL-10}
In experiments of STL-10 classification dataset, a stochastic gradient descent optimizer is used and the learning rate is set as 0.1 during 10 epochs. 
The learning rate is controlled to get diminished by a factor of 0.1 in every additional 30 epochs until 200 epochs.
We have augmented the original images with the size of $96 \times 96$ by adding paddings of 4, randomly cropping them into the size of $96\times96$ again, 
and normalizing RGB channels with the factors of mean, (0.4914, 0.4822, 0.4465), and variance, (0.2023, 0.1994, 0.2010). For ResNet-18 model, since the input image size of STL-10 images is $96 \times 96$,  kernel sizes in the very first convolutional layer are modified from $7 \times 7$ to $3 \times 3$ while max pooling layers are eliminated. 
The same loss function for classification is used as in the experiments on Scaled-MNIST dataset.

\subsubsection{Sort-of-CLEVR}
For sort-of-CLEVR dataset, we have vectorized question data so that we do not have to deal with addtional LSTM training. 
We use original images sized in $75 \times 75$ without performing any augmentations.
The candidate models are optimized with ADAM optimizers with learning rates of 0.001 during first 20 epochs and 0.0001 during next 30 epochs. We have early-stopped the training when the training reaches 50 epochs.

\subsubsection{CLEVR}
Unlike Sort-of-CLEVR dataset, an additional LSTM model that vectorizes questions is also needed to be trained together 
for end-to-end training of CLEVR dataset.
An ADAM optimizer is used with a learning rate of 0.00025 during first 200 epochs and 0.000025 during next 200 epochs. 
And, we use a single GPU with a batch size of 64 instead of the 10-distributed computing method performed in \cite{relationalreasoningmodule}.
Optimizing the LSTM model parameters is a much difficult job without a weight decay factor because of training  instability, and thus weight decay rate of 0.0001 is also applied. 
The LSTM model is structured with two layers of 128 bi-directional hidden units, taking pre-trained GloVe~\cite{glove} embeddings with six billion tokens and 50-dimensional vectors

\section{Additional Experiments}

\setlength{\tabcolsep}{4pt}
\begin{table*}[t]
\centering
\caption{\textbf{Robustness against Deformation on Scaled MNIST.} The results of the networks with dilated convolutions and deformable convolutions may not necessarily represent the performances of their algorithms because the models we have experimented are modularized versions of their proposed concepts}
\label{table:scaled_mnist-bcn}
\resizebox{0.99\linewidth}{!}{
\begin{tabular}{l|c|c|c|c|c}
\hline  
& \multicolumn{5}{c}{ Classification Accuracy } \\ 
\cline{2-6}
& Scaled-MNIST & MoreScaled-MNIST & with $\pm 10^{\circ}$ & with $\pm 45^{\circ}$  & affNIST\\
\hline
Baseline & 85.4\% &  19.0\% & 20.0\% & 19.7\% & 27.9\% \\
Baseline(depth 4) & 95.6\% & 53.4\% & 52.1\% & 41.6\% & 61.0\% \\
Baseline(depth 5) & 96.6\% & 71.3\% & 69.2\% &52.2\% & 77.4\% \\
Baseline(depth 5, 2$\times$ filter) & \textbf{97.5}\% & 88.5\% & 86.9\% & 65.2\% & 78.8\% \\
\hline
Base + deformable Conv \cite{deformconv} & 93.0\% & 27.9\% &28.3\% &25.6\% & 29.2\% \\
Base + dilated Conv \cite{yu2015multi} & 90.5\% &27.0\% &28.0\% &25.3\% &  29.3\% \\
\hline
\textbf{Base + BCN} & 97.2\% & \textbf{94.8}\% & \textbf{92.7}\% & \textbf{76.1}\% & \textbf{91.6\%} \\
\hline
\end{tabular}} 
\end{table*}
\setlength{\tabcolsep}{1.4pt}
\subsection{Robustness against Deformations }
Previously introduced experiments on Scaled-MNIST are performed using the networks trained to minimize the hybrid loss function consisting of two terms:
classification error (softmax classifier with cross-entropy cost) and localization error (distance error from the target center points). 
In order to objectively demonstrate the classification performance of BCN on the \textbf{Scaled-MNIST} dataset, the experiments were repeated with a single classification loss implementation.

While conventional approaches of feature generalization in computer vision tasks include data augmentation, invariant feature design, and other additional hand-crafted engineerings of feature representations, 
these attempts 
are not fully robust against
geometric deformations of visual scenes. 
Globally broadcasting spatial features in all receptive fields, our method achieves spatial invariance against deformations in scales and rotations. 

\begin{figure*}[t]
\centering
\includegraphics[width=.9\linewidth]{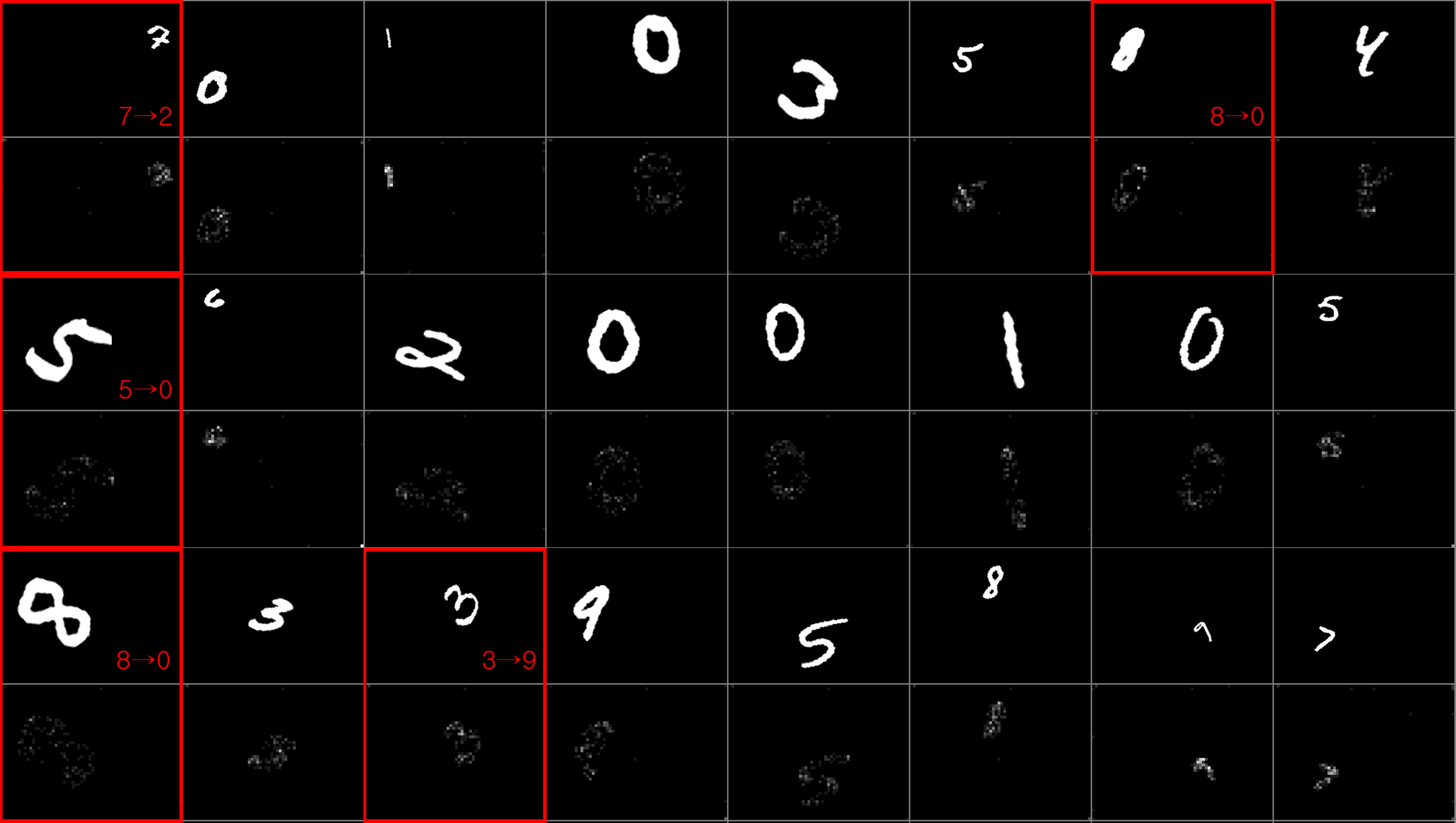}
\caption{Comparison of activation maps for MoreScaled-MNIST with maximum of $\pm 45^{\circ}$ rotation. Red-boxed figures are the misclassified digits with Base+BCN}
\label{fig:mnist_256_big}
\end{figure*}

In a newly generated dataset, \textbf{MoreScaled-MNIST}, we have located the original MNIST dataset (28 $\times$ 28) randomly within 256 $\times$ 192 image space 
while scaling the width of the original digit image from 28 to 160 pixels and the aspect ratio as 1:0.8 to 1:1.2 with uniform randomness.
This dataset has not seen by the models during training, and is used solely for testing. 
All models are optimized through reducing classification loss. 
Additional to the scale deformations, we have tested two cases of rotation ($\pm 10^{\circ}, \pm 45^{\circ} $) applied on each MoreScaled-MNIST image. 
\textbf{affNIST}\footnotemark, lastly, is also tested. 
affNIST consists of original MNIST images with various reasonable affine transformations applied. 
The same models trained for the classification experiment on Scaled-MNIST dataset are tested 
on affNIST, MoreScaled-MNIST and its rotated versions without any additional training and data augmentation.
The results are self-explanatory in Table \ref{table:scaled_mnist-bcn}.
And Figure \ref{fig:mnist_256_big} depicts activation maps showing brocasted information of MoreScaled-MNIST images with random rotations within a range of $\pm45^{\circ}$. Mispredicted images are red-boxed with target labels pointing at predicted labels at the bottom right side. Since the test images are scaled larger and rotated, the model with BCN tends to predict circinate digits as zeros. 

\footnotetext{http://www.cs.toronto.edu/~tijmen/affNIST/}

\subsection{Sectional Runtimes}
For an explicit comparison of computations performance, 
we have measured floating point operations (FLOPs) on sectional parts of RN and multiRN models on the Sort-of-CLEVR experiment.
Applying the stride size of 1 instead of 2 in the CNN is denoted as CNN$_h$ in Table \ref{table:ops} and this application is done for both of RN and multiRN. 
Changing the stride of the last convolution layer results in the change of the feature map size from $5\times5$ to $10\times10$. 
While the MLP operation of $g_\theta$ in CNN$_h$+RN requires 16 times more of floating point operations after increasing the feature map size four times (two for each dimension) larger than CNN+RN, 
the section of convolving the outputs after broadcasting requires just four
times more of operations after increasing feature map size the same amount, which implies linear increments along the size of feature maps. 
Table \ref{table:ops} shows the extravagant computation cost gaps in total FLOPs among different methods. 
It shows vast computational efficiency of our method compared to RN.

\setlength{\tabcolsep}{4pt}
\begin{table}[t]
\centering
\caption{Comparison of the number of FLOPs on Sort-of-CLEVR. Input convolutions and $f_\phi$ are the same, but BCN and different $g_\theta$ is implemented in multiRN }
\label{table:ops}
\begin{tabular}{l|cccc|c}
\hline
         & Input Convolution  & $BCN$ &   $\sum{g_\theta}$  & $f_\phi$ & Total FLOPs \\ \hline
CNN+RN & 3.60M  & - & 133.76M & 133.4K & 137.49M \\ 
CNN$_h$+RN  & 3.98M   &  - &2.26G  & 133.4K & 2.26G  \\ \hline
CNN+multiRN & 3.60M & 3.22M & 1.67M  & 133.4K & 8.62M   \\ 
CNN$_h$+multiRN & 3.98M  & 12.87M & 6.62M  & 133.4K & 23.60M  \\ \hline
\end{tabular}
\end{table}
\setlength{\tabcolsep}{1.4pt}

\subsection{Effect of the size of BCN on MultiRN.}

\setlength{\tabcolsep}{4pt}
\begin{table}[t]
\caption{\textbf{Results for various BCN sizes of mutiRN.} 
This is the result of experiments on Sort-of-CLEVR task by changing the size of BCN in CNN$_h$+multiRN model. Each result is an average of three validation accuracies
}
\label{table:size}
\centering
\resizebox{0.90\linewidth}{!}{
\begin{tabular}{l|c|c|c|c}
\hline
BCN Size & Relational & Non-relational & \#Params & Total FLOPs\\
\hline
w/o BCN & 91.3\% & 99.3\%  & 224K &  12.5M \\
$[64, 64]$ & 95.6\% & 99.9\%  & 246K & 14.1M  \\
$[128, 128]$ & 95.5\% & 99.9\%  & 277K  & 17.2M \\
$[256, 256]$ & 95.9\% & 99.9\%   & 362K &  25.8M \\
$[512, 512]$ & 95.5\% & 99.9\%   & 632K & 52.8M \\
\hline
$[64, 64, 64]$ & 96.0\% & 99.9\% & 250K  &  14.6M \\
$[128, 128, 128]$ & 96.2\% & 99.9\% & 293K  & 18.9M \\
$[256, 256, 256]$ & 95.9\% & 99.9\% & 428K  &  32.4M \\
\hline
$[64, 64, 128]$ & 96.1\% & 99.9\% &  271K  & 16.6M  \\
$\textbf{[128,128,256]}$ & \textbf{96.7\%} & 99.9\%   & 345K & 23.6M \\
\hline 
\end{tabular}
} 
\end{table}
\setlength{\tabcolsep}{1.4pt}
Table 3 shows the results of experiments on the effect of BCN size on mutiRN. We have changed only the size of BCN of the CNN$_h$+multiRN model, and have measured the performance in the Sort-of-CLEVR task. 
The model of 3-layered BCN, each layer of which has $[128, 128, 256]$ number of channels, respectively, shows the best relational accuracy. 
Considering the size of BCN  does not have a significant effect for non-relational queries and removing BCN from multiRN allows the model to reach as high as 99.3\% of accuracy for non-relational queries, 
BCN module can be interpreted to influence its overall network much more for the relational problems than the non-relational ones.

In details, BCN with two layers shows the best performance at $[256, 256]$ size, and BCN with a larger or smaller size yields a little bit lower performance. 
The models using 3-layered BCN show higher performance than those with 2 layers. 
Especially, the $[128,128,128]$ model shows higher performance with smaller parameters and computation than $[256,256]$. 
The models with a bottleneck (more channels in the deeper layer) in 3-layered BCN have achieved high performance while reducing the number of parameters. 
The model with $[128, 128, 256]$ which is used in the paper also has a bottleneck and shows the best performance. 
Considering the structure of BCN, max-pooling in the last layer causes the output dimension to be greatly reduced. 
As the size of the last layer of BCN increases, the output dimension after max-pooling also becomes larger. So more information can be broadcast.

\begin{figure}[h]
\centering
\subfigure[Scaled-MNIST]{\includegraphics[width=.9\linewidth]{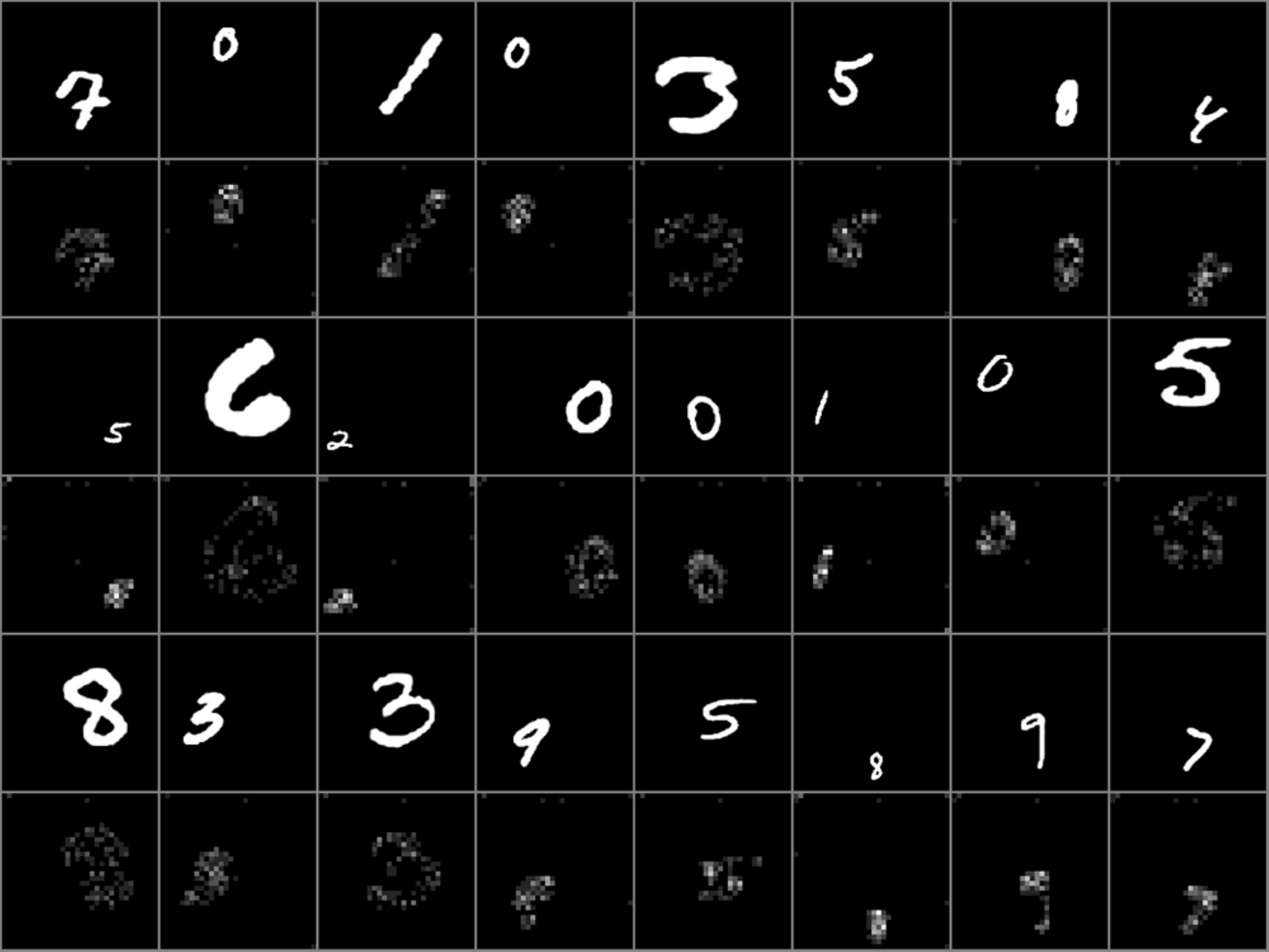}}
\subfigure[STL-10]{\includegraphics[width=.9\linewidth]{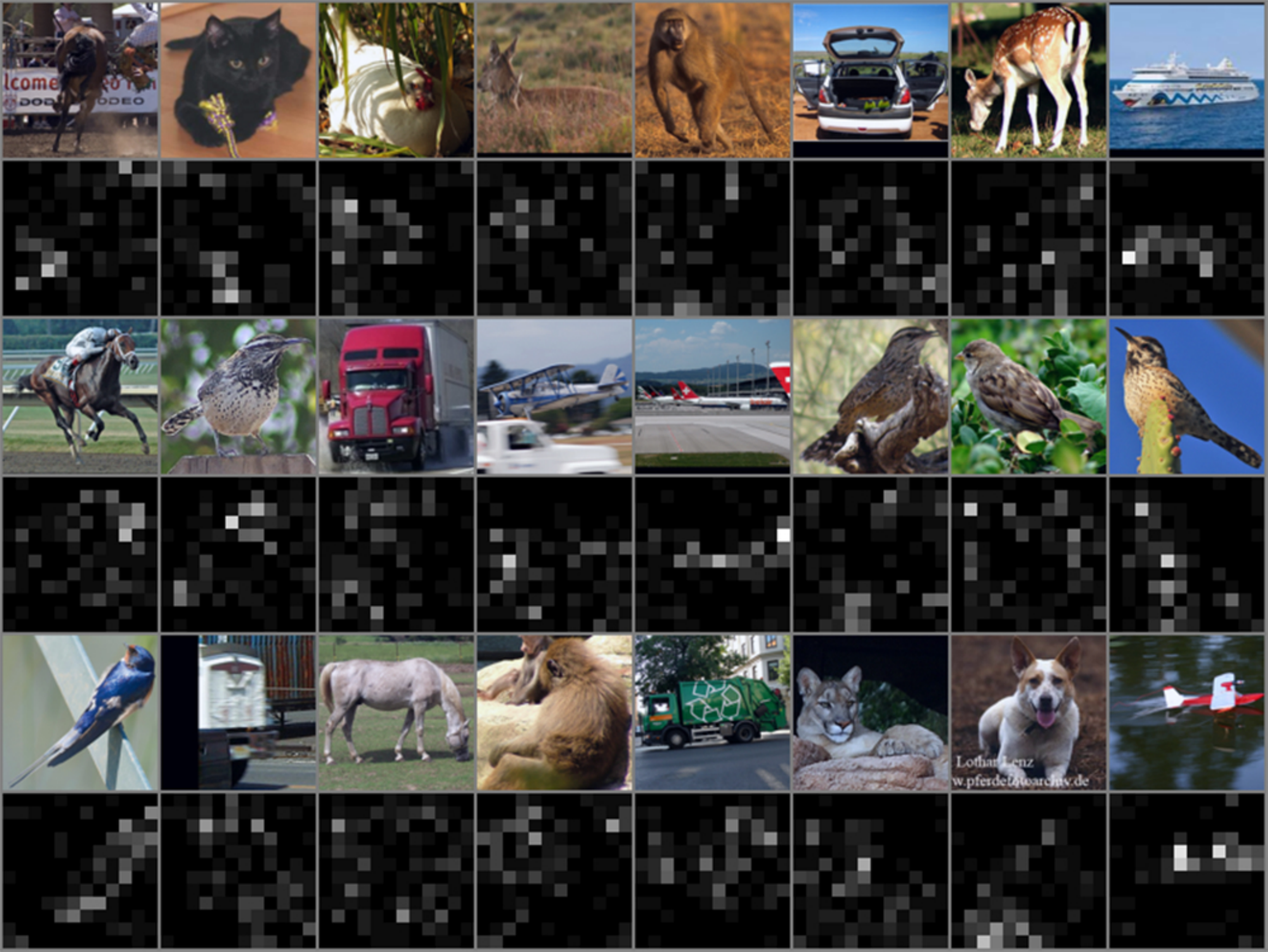}}
\caption{Activation Maps for Image Classification tasks (odd rows: original images, even rows: activation maps). These are the outputs for randomly selected data without manual selection}
\label{fig:act_1}
\end{figure}

\begin{figure}[h]
\centering
\subfigure[Sort-of-CLEVR]{\includegraphics[width=.9\linewidth]{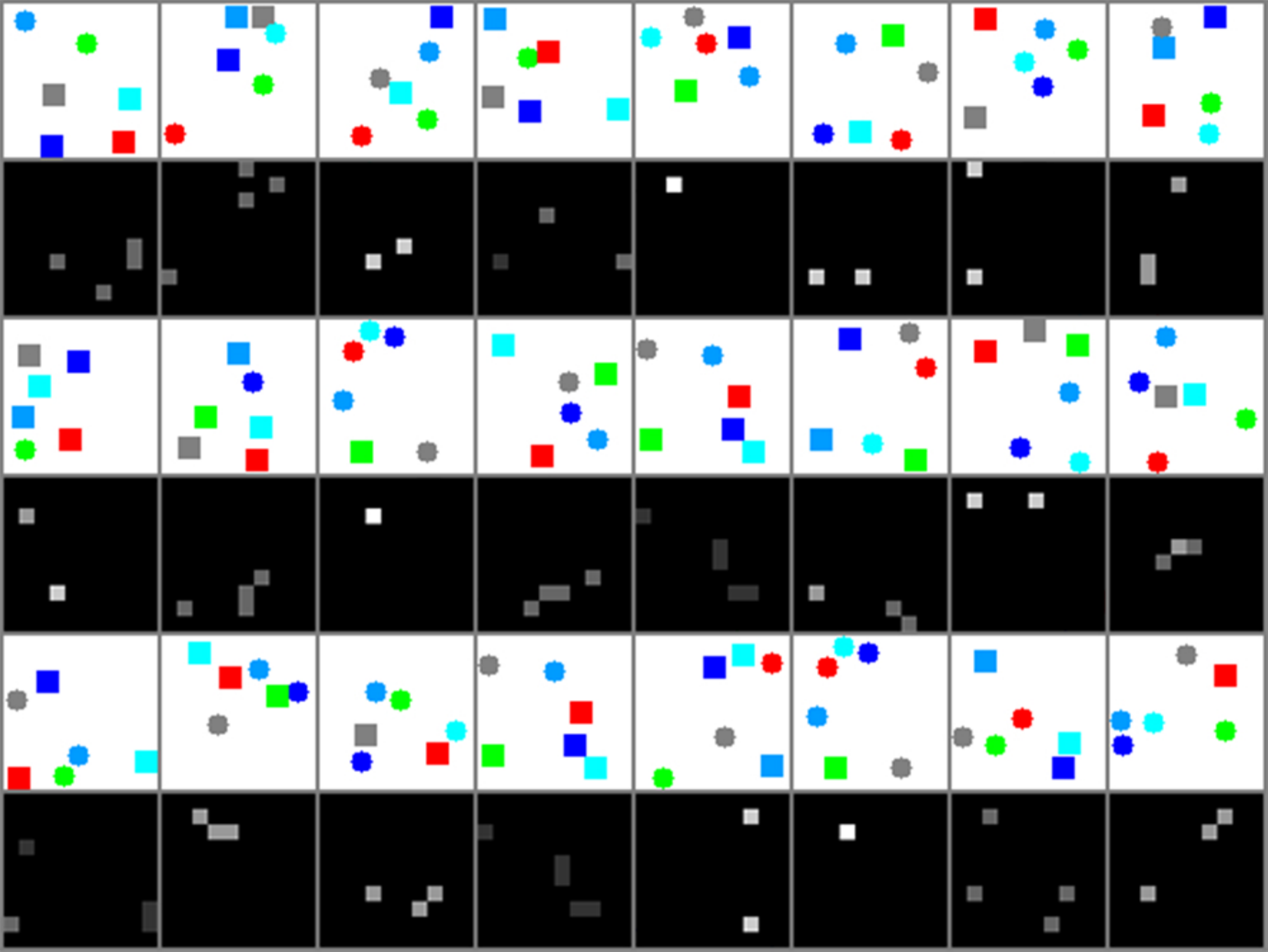}}
\subfigure[CLEVR]{\includegraphics[width=.9\linewidth]{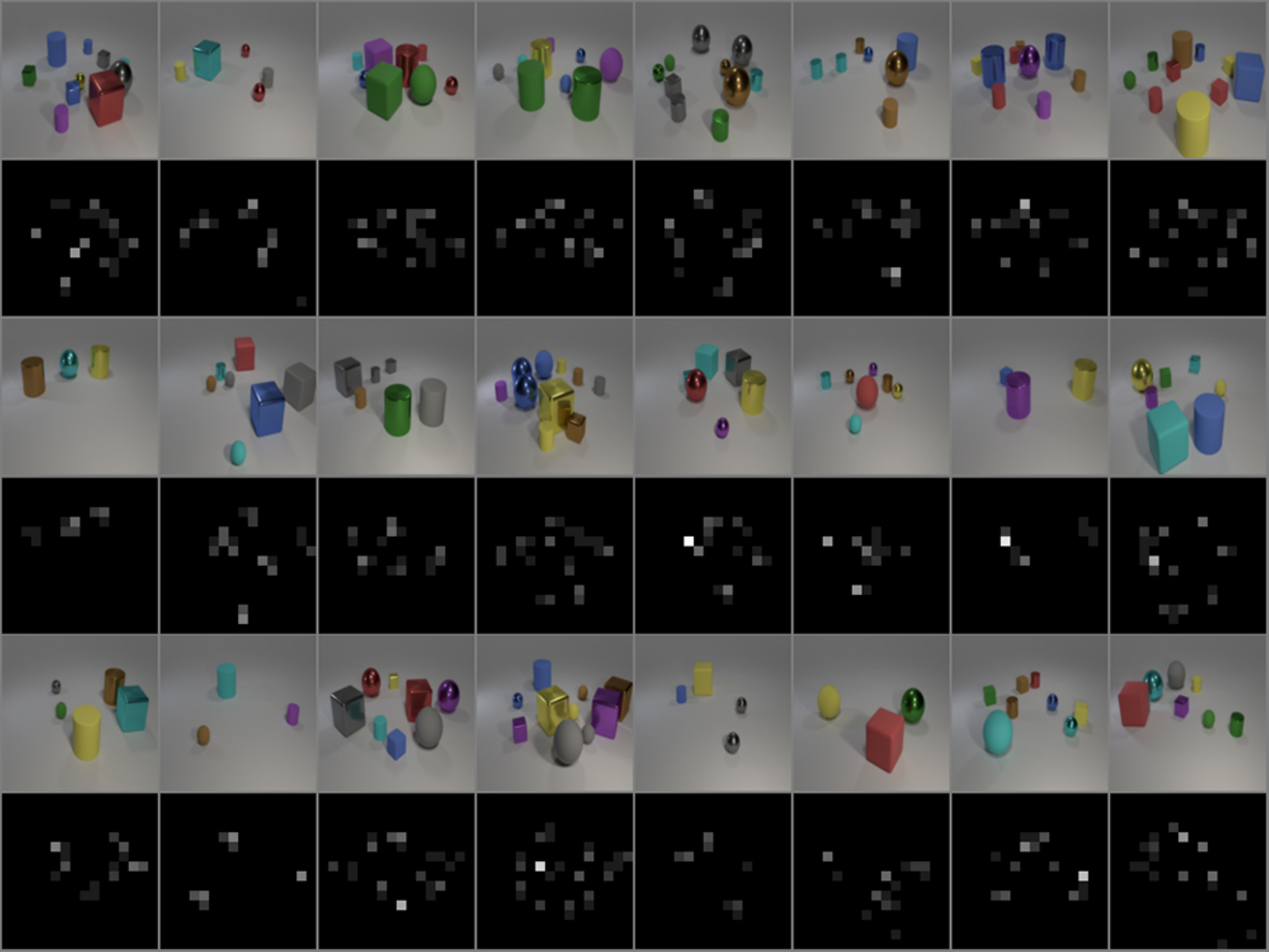}}
\caption{Activation Maps for Relational Reasoning tasks (odd rows: original images, even rows: activation maps). These are the outputs for randomly selected data without manual selection}
\label{fig:act_2}
\end{figure}

\subsection{More Activation Maps}
Figure \ref{fig:act_1} and Figure \ref{fig:act_2} are additional activation maps for the tasks tested in the paper. 
These are randomly selected samples to present how BCN learns to broadcast effective spatial features. 
The brighter parts indicate spatial locations where more information is broadcast.

\end{document}